\documentclass[twoside,leqno,twocolumn]{article}  
\usepackage{ltexpprt} 
\usepackage{graphicx, graphics}
\usepackage{authblk}
\usepackage{caption}
\usepackage{subcaption}
\usepackage{mathrsfs}
\usepackage{fullpage}
\usepackage{vmargin}
\usepackage{varwidth}

\newcommand{\ben}{\begin{enumerate}}
\newcommand{\een}{\end{enumerate}}

\newcommand{\bo}{\textbf}

\newcommand{\x}{\textbf{x}}
\newcommand{\bP}{\textbf{P}}

\newcommand{\bS}{\textbf{S}}

\newcommand{\X}{\textbf{X}}

\newcommand{\e}{\textbf{e}}

\begin{document}

\title{Determining the Number of Clusters via Iterative Consensus Clustering \thanks{This research was supported in part by grants DMS-1127914, NSA REU H9823-10-1-0252, NSF REU DMS-01063010}}
\author{Carl Meyer\thanks{North Carolina State University, Mathematics, Institute of Advanced Analytics \& SAMSI (meyer@ncsu.edu)  }\qquad
Shaina Race \thanks{North Carolina State University, Mathematics \& Operations Research (slrace@ncsu.edu)}\qquad
Kevin Valakuzhy \thanks{University of North Carolina Chapel Hill,\\ Mathematics (kevin.valakuzhy@gmail.com)}}
\maketitle

\begin{abstract}
We use a cluster ensemble to determine the number of clusters, $k$, in a group of data. A consensus similarity matrix is formed from the ensemble using multiple algorithms and several values for $k$. A random walk is induced on the graph defined by the consensus matrix and the eigenvalues of the associated transition probability matrix are used to determine the number of clusters. For noisy or high-dimensional data, an iterative technique is presented to refine this consensus matrix in way that encourages a block-diagonal form. It is shown that the resulting consensus matrix is generally superior to existing similarity matrices for this type of spectral analysis.
\end{abstract}
\section{Introduction}
Ensemble Methods have been used in various data mining fields to improve the performance of a single algorithm or to combine the results of several algorithms. In data clustering, these same strategies have been implemented, and the techniques are commonly referred to as consensus methods \cite{consensusspectral,monti}. Since no single algorithm will work best in any given class of data, it is a natural approach to use several algorithms to solve clustering problems. However, the vast majority of clustering algorithms in the literature require the user to specify the number of clusters, $k$, for the algorithm to create. In applied data mining, the problem is that it is unusual for the user to know this information before hand. In fact, the number of distinct groups in the data may be the very question that the data miner is attempting to answer. \par
This paper proposes a solution to this fundamental problem by using multiple algorithms with multiple values for $k$ to determine the most appropriate value for the number of clusters. 
We begin with a brief theoretical motivation and an example which provides the intuition behind our basic approach. We will follow this discussion with results on real datasets which demonstrate the effectiveness of our iterated approach.
\subsection{Data}
\label{sec:background}
 Let $\X=[\bo{x}_1,\bo{x}_2,\dots,\bo{x}_n]$ be an $m\times n$ matrix of column data.  For the particular implementation of our iterated consensus clustering (ICC) approach outlined herein, we assume the data in $\X$ is nonnegative and relatively noisy. Neither of these conditions are necessary for the general scheme of ICC but one of our preferred algorithms for dimension reduction is nonnegative matrix factorization (NMF), which, as the name suggests requires nonnegative data. Our main focus falls in the realm of document clustering, but we demonstrate that our method works equally well on other types of data. In document clustering the data matrix $\X$ is a term-by-document matrix where $\X_{ij}$ represents the frequency of term $i$ in document $j$. The data in $\X$ are normalized and weighted according to term-weighting schemes like those described in \cite{berryCIR,termweighting,kogan}.
\subsection{Similarity~Matrices}\qquad
A similarity\hfill\break
 matrix $\bS$ is an $n\times n$ symmetric matrix of pairwise similarities for the data in $\X$, where $\bS_{ij}$ measures some notion of similarity between $\x_i$ and $\x_j$. Many clustering algorithms, particularly those of the spectral variety rely on a similarity matrix to cluster data points \cite{shi,ng,spectraltutorial,poweriteration,meila}.  While many types of similarity functions exist, the most commonly used function in the literature is the Gaussian similarity function, $\bS_{ij}=\mbox{exp}(-\frac{\|x_i-x_j\|}{2\sigma^2})$, where $\sigma$ is a parameter, set by the user. We will discuss our own similarity matrix, known as a consensus matrix, in Section \ref{sec:consensusmatrix}. The goal of clustering is to create clusters of objects that have high intra-cluster similarity and low inter-cluster similarity. Thus any similarity matrix, once rows and columns are ordered by cluster, should have a nearly block-diagonal structure.

\subsection{Nearly Uncoupled Markov Chains}

\begin{figure}[h!]

\begin{subfigure}[h]{\linewidth}
\centering
\includegraphics[scale=.22]{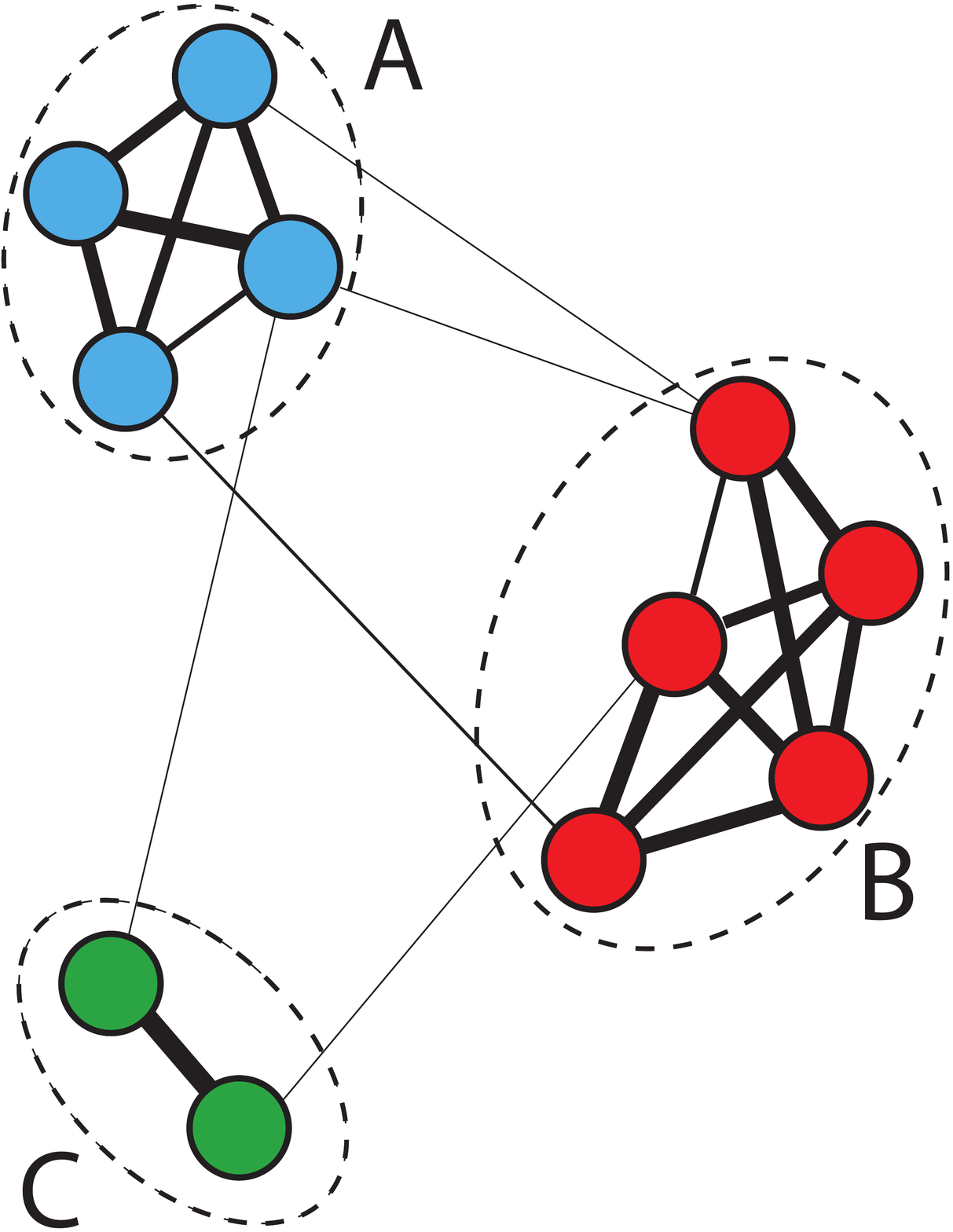}
\end{subfigure}

\begin{subfigure}[h]{0.4\linewidth}
\centering
\includegraphics[scale=.3]{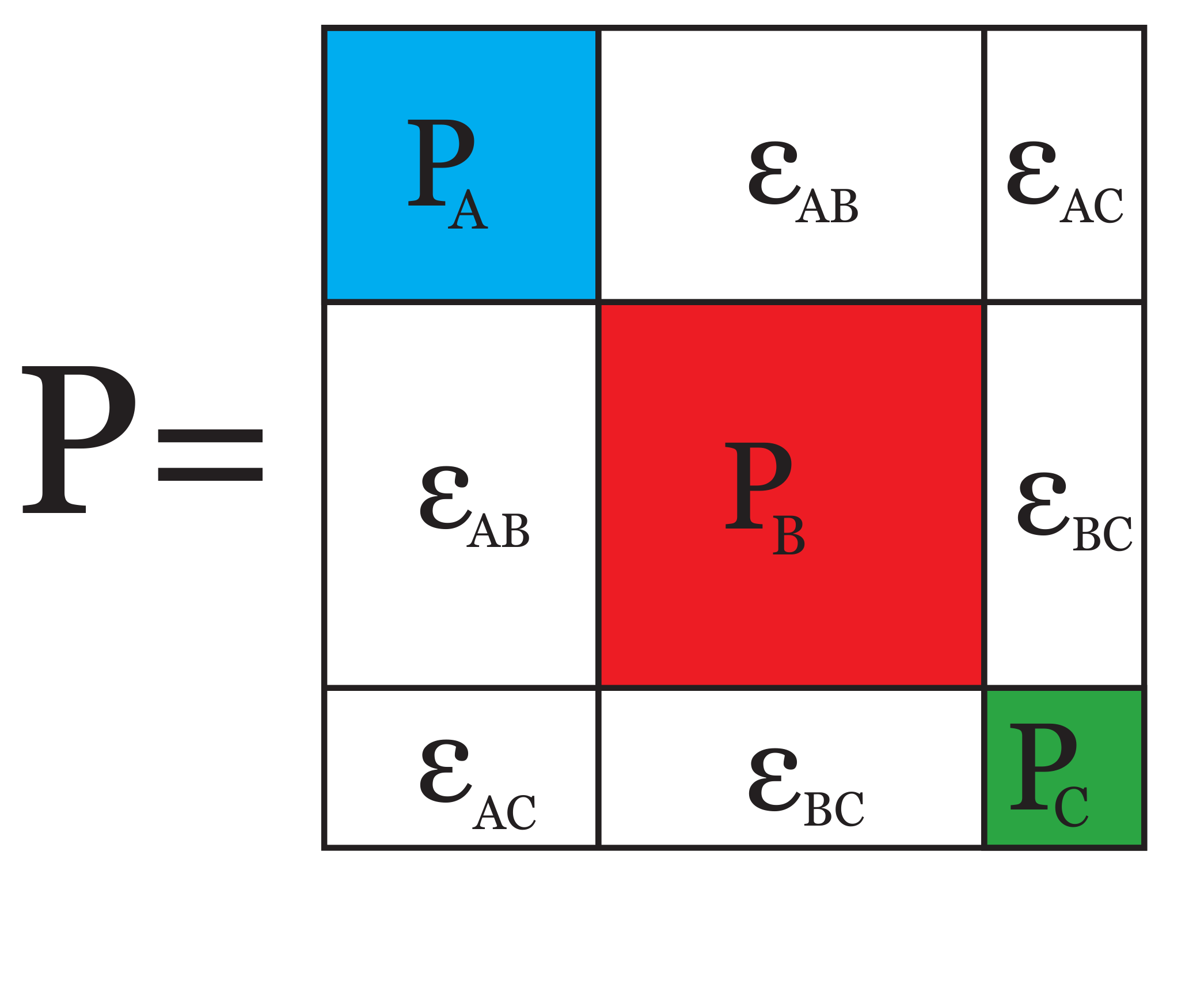}
\end{subfigure}
\caption{A Nearly Uncoupled Markov Chain}
 \label{fig:NuMC}
\end{figure}

Any similarity matrix, $\bS$, can be viewed as an adjacency matrix for nodes on an undirected graph. The $n$ data points act as nodes on the graph and edges are drawn between nodes with weights from the similarity matrix. Figure \ref{fig:NuMC} illustrates such a graph, using the thickness of an edge to indicate its weight. While edges may exist between nodes in separate clusters, we expect the weights of such edges to be far less than the weights within the clusters. 

A random walk is induced on the graph and a transition probability matrix, $\bP$, is created from the similarity matrix, $\bo{S}$, as $\bP=\bo{D}^{-1}\bo{S}$ where $\bo{D}=diag(\bo{S}\e)$, and $\e$ is a vector of ones. It is easily verified that a steady-state distribution of this Markov chain is given by $\pi^{T} = \frac{\e^T\bo{D}}{\e^T\bo{D}\e}$.  Let $\bo{Q}=diag(\pi)=\frac{\bo{D}}{\e^T\bo{D} \e}$.  $\bP$ represents a \textit{reversible} Markov chain because it satisfies the detailed balance equations, $\bo{Q}\bP=\bP^T \bo{Q}$ \cite{kemenysnell,stewart}. This condition guarantees that the eigenvalues of $\bP$ are real since $\bo{Q}^{1/2}\bP \bo{Q}^{-1/2}=\bo{Q}^{-1/2}\bP^T \bo{Q}^{1/2}$ indicates that $\bP$ is similar to a symmetric matrix. In fact, this symmetric matrix, $\bo{Q}^{1/2}\bP \bo{Q}^{-1/2}$, is precisely $\bo{I}-\mathscr{L}$ where $\mathscr{L}$ is the \bo{normalized Laplacian} matrix used in many spectral clustering algorithms \cite{spectraltutorial}. For computational considerations, we use this symmetric matrix to compute the spectrum of $\bP$ in our algorithm. \par

Let $\sigma(\bP)=\{1=\lambda_1 \geq \lambda_2 \geq \dots \geq \lambda_n\}$ be the spectrum of $\bP$. A block diagonally dominant Markov Chain is said to be \textit{nearly uncoupled} if the diagonal blocks of $\bP$ are themselves \textit{nearly stochastic}, meaning $\bP_i \e \approx 1$ for each $i$ (for a more precise definition, see \cite{chuck}). A nearly uncoupled Markov chain with real eigenvalues will have exactly $k$ eigenvalues near $1$ where $k$ is the number of blocks on the diagonal.  This cluster of eigenvalues, $[\lambda_1,\dots,\lambda_k]$, near $1$ is known as the \bo{Perron cluster} \cite{fischer, chuck, chuckthesis}. Moreover, if there is no further decomposition (or meaningful sub-clustering) of the diagonal blocks, a relatively large gap between the eigenvalues $\lambda_k$ and $\lambda_{k+1}$ is expected \cite{fischer,chuck,chuckthesis}. It has previously been suggested that this gap be observed to determine the number of clusters in data \cite{spectraltutorial}. However, as we will demonstrate in Section \ref{sec:results}, the most common similarity matrices in the literature do not impart the level of uncoupling that is necessary for a visible Perron cluster.
  The main goal of our algorithm is to construct a nearly uncoupled Markov chain using a similarity matrix from a cluster ensemble. The next section fully motivates our approach.

\section{The Consensus Similarity Matrix}
\label{sec:consensusmatrix}
We will build a similarity matrix using the results of several, say $N$, different clustering algorithms. As previously mentioned, most clustering algorithms require the user to input the number of desired clusters. We will choose 1 or more values for $k$, denoted by $\tilde{k} = [\tilde{k}_1, \tilde{k}_2, \dots, \tilde{k}_J]$, and use each of the $N$ algorithms to partition the data into $\tilde{k}_i$ clusters, for $i=1,\dots,J$. The result is a set of $JN$ clusterings. These clusterings are recorded in a \textbf{consensus matrix}, $\bo{M}$, by setting $M_{ij}$ equal to the number of times observation $i$ was clustered with observation $j$.  Such a matrix has become popular for ensemble methods, see for example \cite{monti,consensusspectral}. We will then observe the eigenvalues of the transition probability matrix of the random walk on the graph associated with the consensus matrix.
\par
\begin{figure}[h!]
\centering
\begin{subfigure}{\linewidth}
  \centering
  \includegraphics[scale=.17]{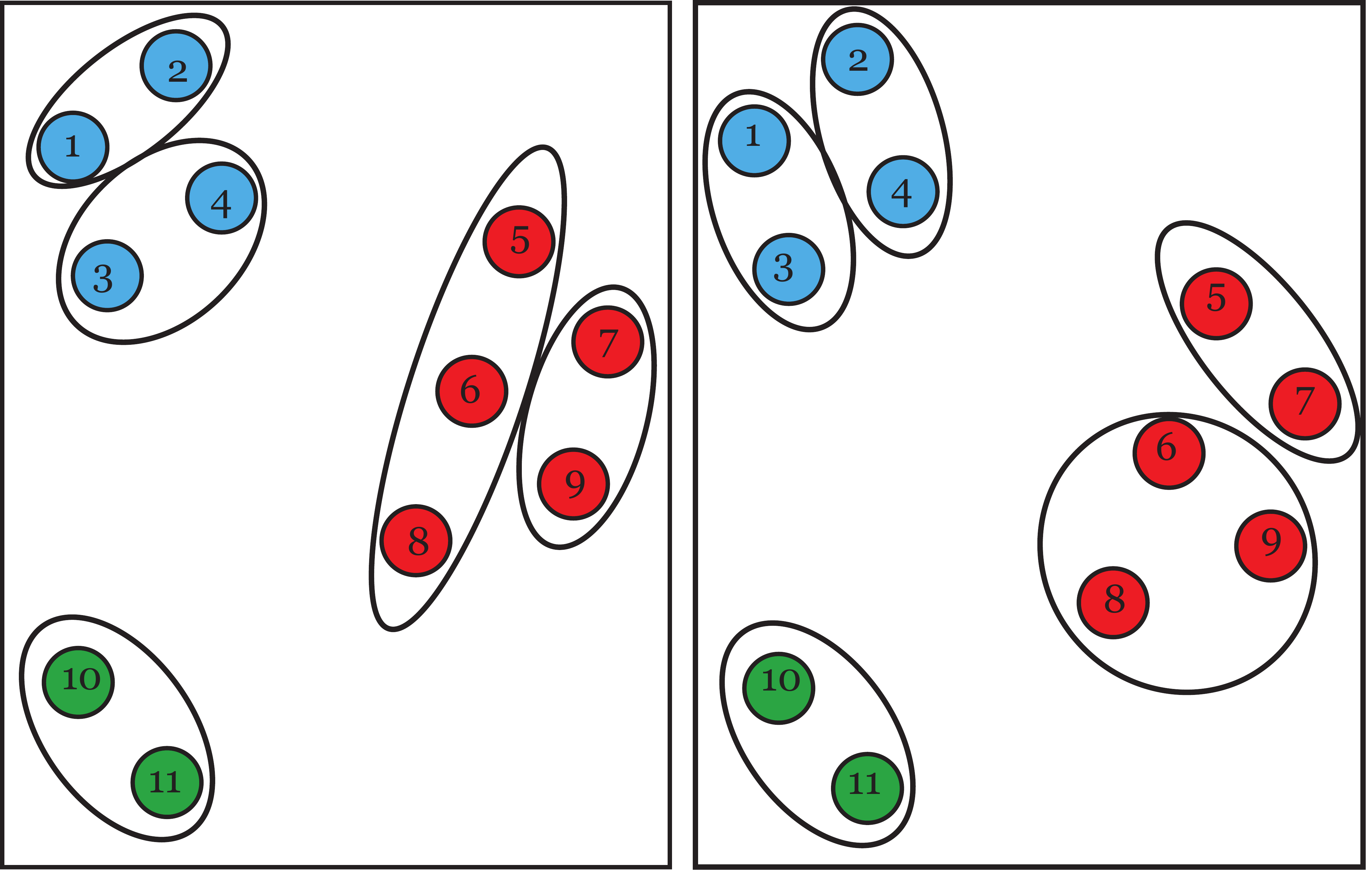}
  \caption{Two Different Clusterings with $\tilde{k}=5$}
  
\end{subfigure}%

\begin{subfigure}{\linewidth}
  \centering
  \includegraphics[scale=.85]{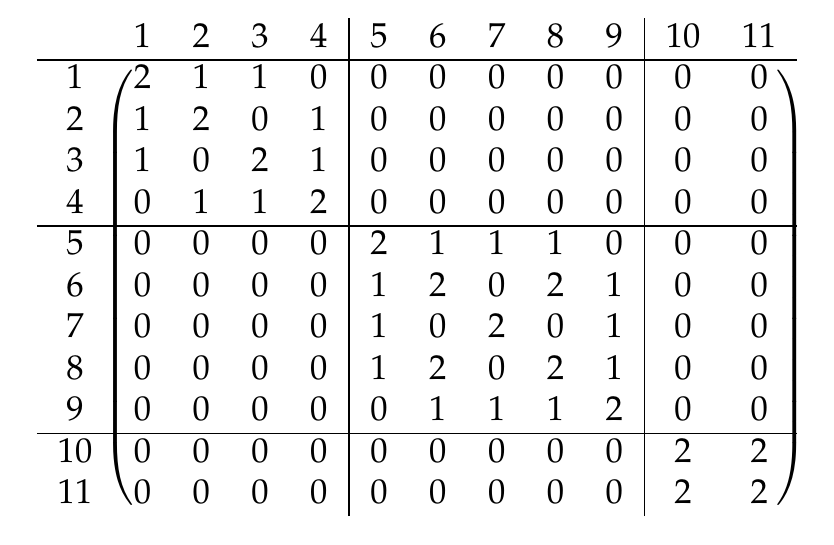}
  \caption{Resulting Consensus Matrix}

\end{subfigure}
\begin{subfigure}{\linewidth}
  \centering
  \includegraphics[scale=.35]{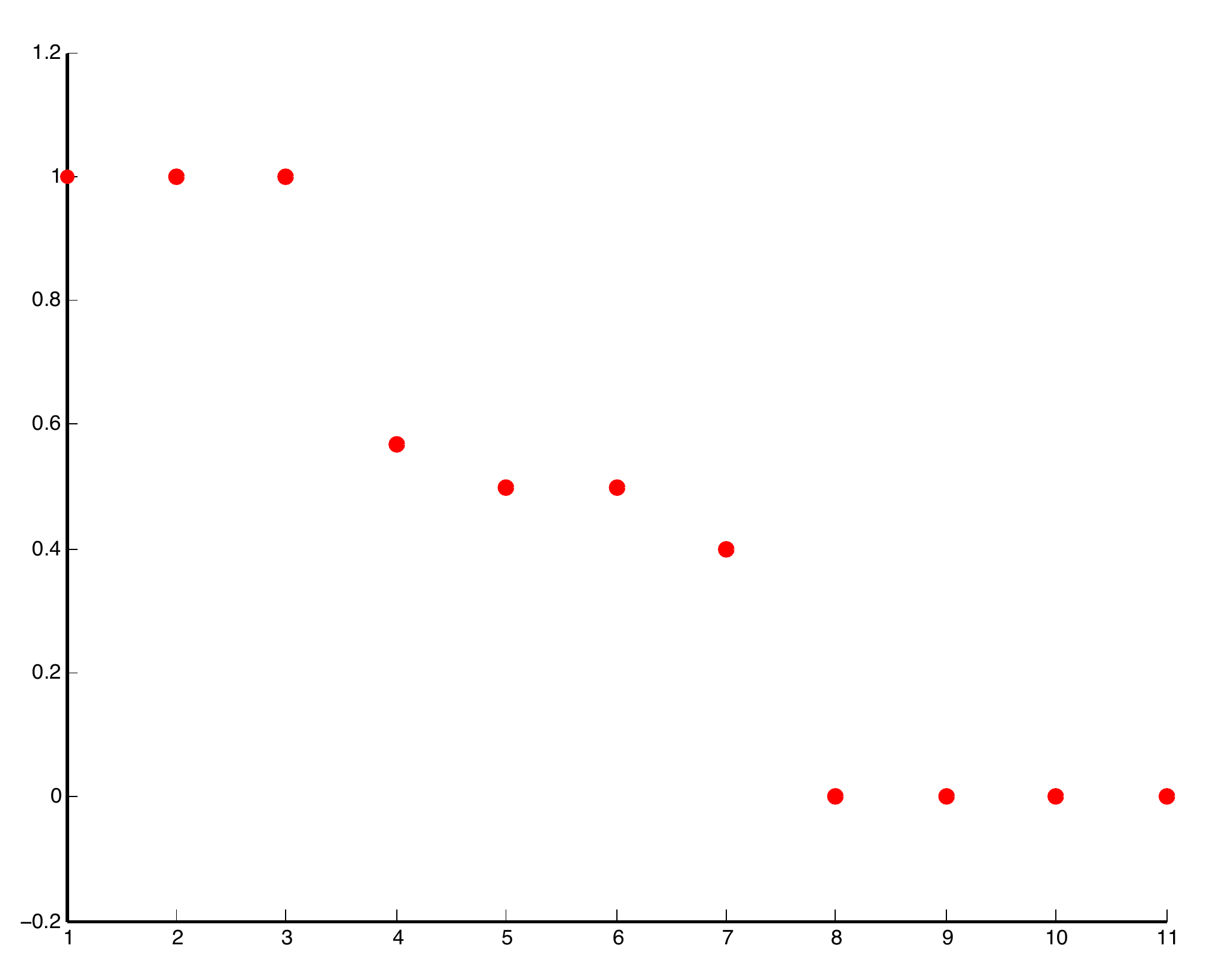}
\caption{$k=3$ Eigenvalues in Perron Cluster}
\end{subfigure}
\caption{A Simple Motivating Example}
\label{fig:example}
\end{figure}

 To motivate our approach, we'll look at a brief fabricated example. We will use the vertices from the graph in Figure \ref{fig:NuMC}  which are clearly separated into 3 clusters. Figure \ref{fig:example} illustrates (a) two different clusterings of these points (each with $\tilde{k}=5$ clusters), (b) the consensus similarity matrix resulting from these two clusterings, and (c) the first few eigenvalues of the transition probability matrix, sorted by magnitude. Using an incorrect guess of $\tilde{k}=5$ and 2 clusterings, the correct value of $k$ is discovered by counting the number of eigenvalues in the Perron cluster.

Our use of this consensus similarity matrix relies on the following assumptions about our underlying clustering algorithms:
 \begin{itemize}
\item If there are truly $k$ distinct clusters in a given dataset, and a clustering algorithm is set to find $\tilde{k}>k$ clusters, then the original $k$ clusters will be broken apart into smaller clusters to make $\tilde{k}$ total clusters.
\item Further, if there is no clear ``subcluster" structure, meaning the original $k$ clusters do not further break down into meaningful components, then different algorithms will break the clusters apart in different ways. 
\end{itemize}  
Before discussing adjustments made to this basic approach, we provide a brief description of the clustering algorithms used herein.
\section{Clustering Algorithms}
\label{sec:algorithms}
The authors have chosen four different algorithms to form the consensus matrix: principal direction divisive partitioning (PDDP) \cite{pddp}, $k$-means, and expectation-maximization with Gaussian mixtures (EMGM) \cite{datamining}. For each round of clustering, $k$-means is run twice, once initialized randomly and once initialized with the centroids of the clusters found by PDDP. This latter hybrid, ``PDDP-$k$-means'', is considered the $4^{th}$ algorithm. For text data sets and the clustering of symmetric matrices, spherical $k$-means is used as opposed to Euclidean $k$-means. \paragraph*{•}
Three different dimension reductions are used to alter the data input to each of these algorithms. Our motivation for this is focused along three objectives. The first objective is merely to reduce the size of the data matrix, which speeds the computation time of the clustering algorithms. The second objective is to reduce noise in the data. The final objective is to decompose the data into components that reveal underlying patterns or features. The first dimension reduction is the ever popular Principal Components Analysis (PCA) \cite{PCA}. The second dimension reduction is a simple truncated Singular Value Decomposition (SVD) \cite{meyerbook}. We use Larson's PROPACK software to efficiently compute both the SVD and PCA \cite{propackpaper}. The third dimension reduction is a nonnegative matrix factorization (NMF) \cite{leeseung,AppliNMF}. The NMF algorithm used is the alternating constrained least squares (ACLS) algorithm \cite{SAS} with sparsity parameters $\lambda_W=\lambda_H=0.5$, and initialization of factor W with the Acol approach outlined in \cite{SAS}. For further explanation on how and why these techniques are used for dimension reduction, see the complete discussion in \cite{phdthesis}. 
 \par
All of the above dimension reduction techniques require the user to input the level of the dimension reduction, $r$.  The choices for this parameter can provide hundreds of different clusterings for a single algorithm. Here, we choose three different values for $r$: $r_1, r_2, r_3$.  For smaller datasets where it is feasible to compute the complete SVD/PCA of the data matrix,  $r_1, r_2, r_3$ were chosen to be the number of principal components required to capture 60\%, 75\% and 90\% of the variance in the data respectively. We require that the values of $r_1,r_2,\,\mbox{and}\,\,\, r_3$ be unique. For larger document datasets ($n\geq 3000$ documents), where it is unwieldy to compute the entire SVD of the data matrix, values for $r$ are chosen such that $r_1\approx 0.01n$, $r_2 \approx 0.05n$, and $r_3 \approx 0.1n$. 
\par
Using our four different algorithms, 4 representations of the data (raw data plus three dimension reductions), three ranks of dimension reduction we can create up to $N=40$ different clusterings for each value of $\tilde{k}$.
 \section{Our Method}
The base version of our method is simple and works well on datasets with well-defined, well-separated clusters. Section \ref{sec:adjustments} discusses enhancements that provide an exploratory method for larger, noisier datasets.
\begin{algorithm}[Basic Method]
\vspace{.01cm}
\begin{itemize}
\item[] \textbf{Input:} Data Matrix $\X$ and a sequence $\tilde{k}=\tilde{k}_1,\tilde{k}_2,\dots,\tilde{k}_J$
\item[1.] Using each clustering method $i=1,\dots,N$, partition the data into $\tilde{k}_j$ clusters, $j=1,\dots,J$
\item[2.] Form a consensus matrix, $\bo{M}$ with the $JN$ different clusterings determined in step 1. Let $\bo{D}=\mbox{diag} (\bo{M}\e$)
\item[3.] Compute the eigenvalues of $\bP$ using the symmetric matrix $\bo{I}-\bo{D}^{-1/2}\bo{M}\bo{D}^{-1/2}$ and identify the Perron cluster.
\item[] \textbf{Output:} The number of eigenvalues, $k$, contained in the Perron cluster.
\end{itemize}
\end{algorithm}

\begin{figure}[ht!]
\centering
  \begin{subfigure}[h]{\linewidth}
  \centering
  \includegraphics[scale=.33]{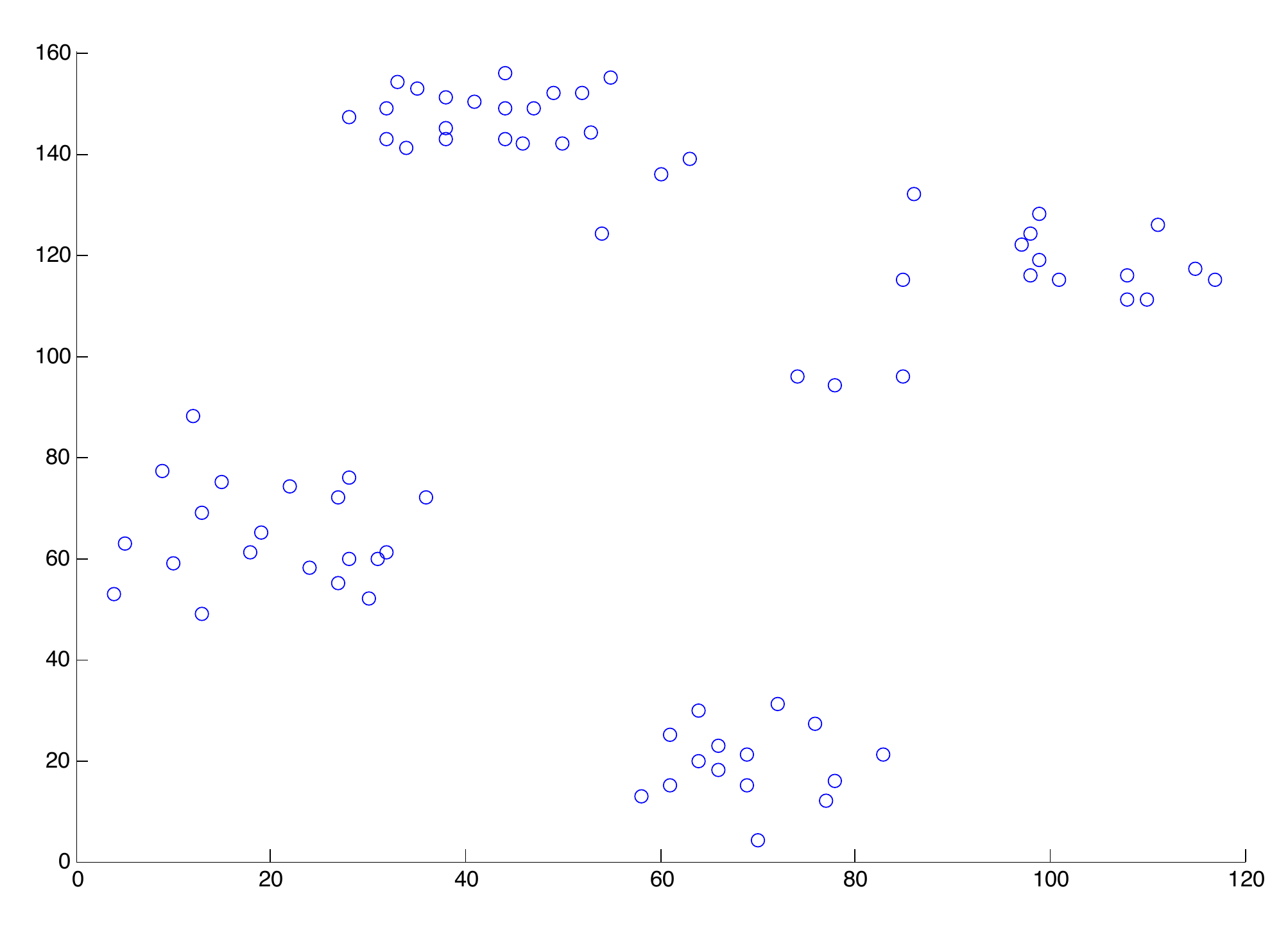}
  \caption{\small Scatter Plot of Ruspini Data}
  \end{subfigure}

  \begin{subfigure}[h]{\linewidth}
  \centering
  \includegraphics[scale=.22]{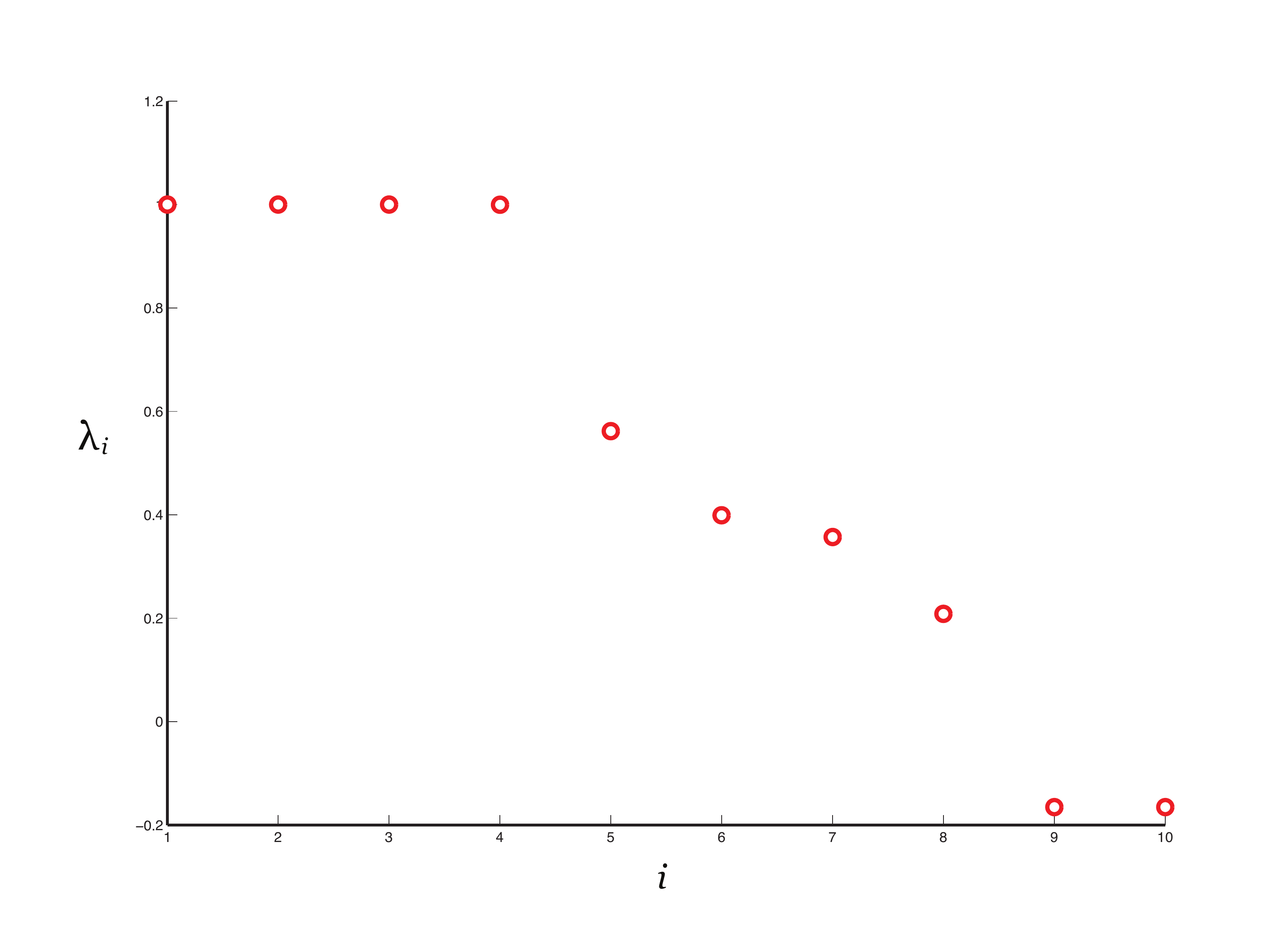}
  \caption{\small Eigenvalues of Probability Matrix}
\end{subfigure}

\caption{Results on Ruspini Dataset}
\label{fig:ruspini}
\end{figure}

To demonstrate the effectiveness of our base method on a simple synthetic dataset, we employed it on the Ruspini dataset, a two-dimensional dataset that has commonly been used to validate clustering methods and metrics. We simply used our four different algorithms and five different values for $\tilde{k}=6,\dots,10$. The resulting eigenvalue plot, displayed in Figure \ref{fig:ruspini}, clearly shows the correct number, $k=4$, of eigenvalues in the Perron cluster. 
\par
\begin{figure}[h!]
\centering
 \includegraphics[scale=.37]{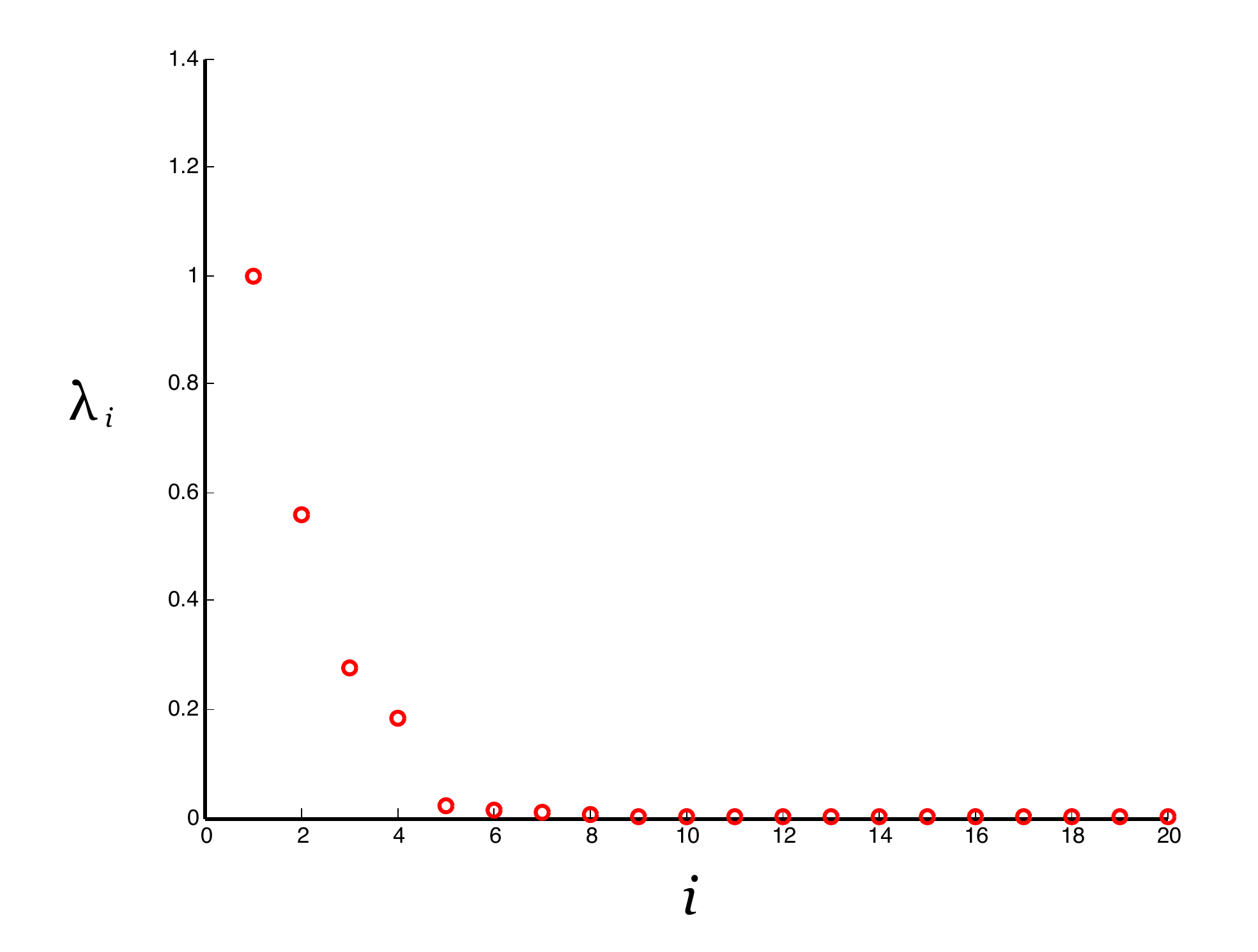}
 \caption{\small Eigenvalues from Gaussian Similarity Matrix for Ruspini Dataset}
 \label{fig:ruspinigaussian}
\end{figure}

 For the purposes of comparison Figure \ref{fig:ruspinigaussian} shows the eigenvalues of the Markov chain induced by the Gaussian similarity matrix. In all of our experiments we set the parameter $\sigma^2=\frac{1}{n-1} \sum_{i=1}^n \|\bo{x}_i-\mu\|_2^2$ where $\mu = \frac{\X\e}{n}$ is the mean. While there are 4 relatively large eigengaps in Figure \ref{fig:ruspinigaussian}, the largest gap occurs after the first eigenvalue and there is little indication of the block-diagonal dominance (uncoupling) illustrated in Figure \ref{fig:NuMC}.

\section{Adjustments to the Algorithm}
\label{sec:adjustments}
This section discusses two adjustments to our algorithm, each of which are meant to combat the effect of noise in large datasets, particularly document sets. In document clustering, although the underlying topics that define individual clusters may be quite distinct, the spatial concept of ``well-separated'' clusters becomes convoluted in high dimensions. Thus the nearly uncoupled structure depicted in Figure \ref{fig:NuMC} is rare in practice. The adjustments presented in this section are meant to refine the data in an iterative way to encourage such uncoupling. 
\subsection{Drop Tolerance, $\tau$} There will necessarily be some similarity between documents from different clusters. As a result,  clustering algorithms make errors. However, if the algorithms are independent, it is reasonable to expect that the majority of algorithms will not make the \textit{same} error. To this end, we introduce a \textit{drop tolerance}, $\tau$, $0\leq \tau <0.5$ for which we will drop (set to zero) entries $\bo{M}_{ij}$ in the consensus matrix if $\bo{M}_{ij}<\tau JN$. For example $\tau=0.1$ means that when $x_i$ and $x_j$ are clustered together in fewer than 10\% of the clusterings they are disconnected in the graph.
\subsection{Iteration} In the basic algorithm previously outlined, we use several clusterings to transform our original data matrix, $\X$, into a similarity matrix, $\bo{M}$. The rows/columns of $\bo{M}$ are essentially a new set of variables describing our original observations. Thus $\bo{M}$ can be used as the data input to our clustering algorithms, and the procedure can be iterated as follows:
\begin{algorithm}[Iterated Method (ICC)]
\vspace{.01cm}
\begin{itemize}
\item[] \textbf{Input:} Data Matrix $\X$, drop-tolerance $\tau$, and sequence $\tilde{k}=\tilde{k}_1,\tilde{k}_2,\dots,\tilde{k}_J$
\item[1.] Using each clustering method $i=1,\dots,N$, partition the data into $\tilde{k}_j$ clusters, $j=1,\dots,J$
\item[2.] Form a consensus matrix, $\bo{M}$ with the $JN$ different clusterings determined in step 1. 
\item[3.] Set  $\bo{M}_{ij}=0$ if $\bo{M}_{ij} <\tau JN$.
\item[4.]Let $\bo{D}=\mbox{diag} (\bo{M}\e)$. Compute the eigenvalues of $\bP$ using the symmetric matrix $\bo{I}-\bo{D}^{-1/2}\bo{M}\bo{D}^{-1/2}$.
\item[5.] If the Perron Cluster is clearly visible, stop and output the number of eigenvalues in the Perron cluster, $k$. Otherwise, repeat steps 1-5 using $\bo{M}$ as the data input in place of $\X$.

\end{itemize}
\end{algorithm}
While the uncoupling benefit of the drop tolerance should be clear from the graph in Figure \ref{fig:NuMC}, the benefit of iteration may not be apparent to the reader until the result is visualized. In the next section, we will use noisy datasets to illustrate.

\section{Results on Noisy Data}
\label{sec:results} In order to demonstrate the uncoupling effect of iteration, we use three datasets that are difficult to cluster because of their inherent noise.
\subsection{Newsgroups Dataset} Our Newsgroups dataset is a subset of 700 documents, 100 from each of $k=7$ clusters, from the 20 Newsgroups dataset \cite{20ng}. The topic labels from which the documents were drawn can be found in Table \ref{tab:newsgroupslabels}.
\begin{table}[h!]
\centering
    \begin{tabular}{c}
    \hline
        Alt: Atheism              \\ 
        Comp: Graphics            \\ 
        Comp: OS MS Windows Misc. \\ 
        Rec: Sport Baseball       \\ 
        Sci: Medicine             \\ 
        Talk: Politics Guns       \\ 
        Talk: Religion Misc.      \\
\hline
    \end{tabular}
\caption{Topics for Newsgroups Dataset}
\label{tab:newsgroupslabels}
\end{table}
\par
\begin{figure}[h!]
\begin{subfigure}[h!]{\linewidth}
\centering
\includegraphics[scale=.35]{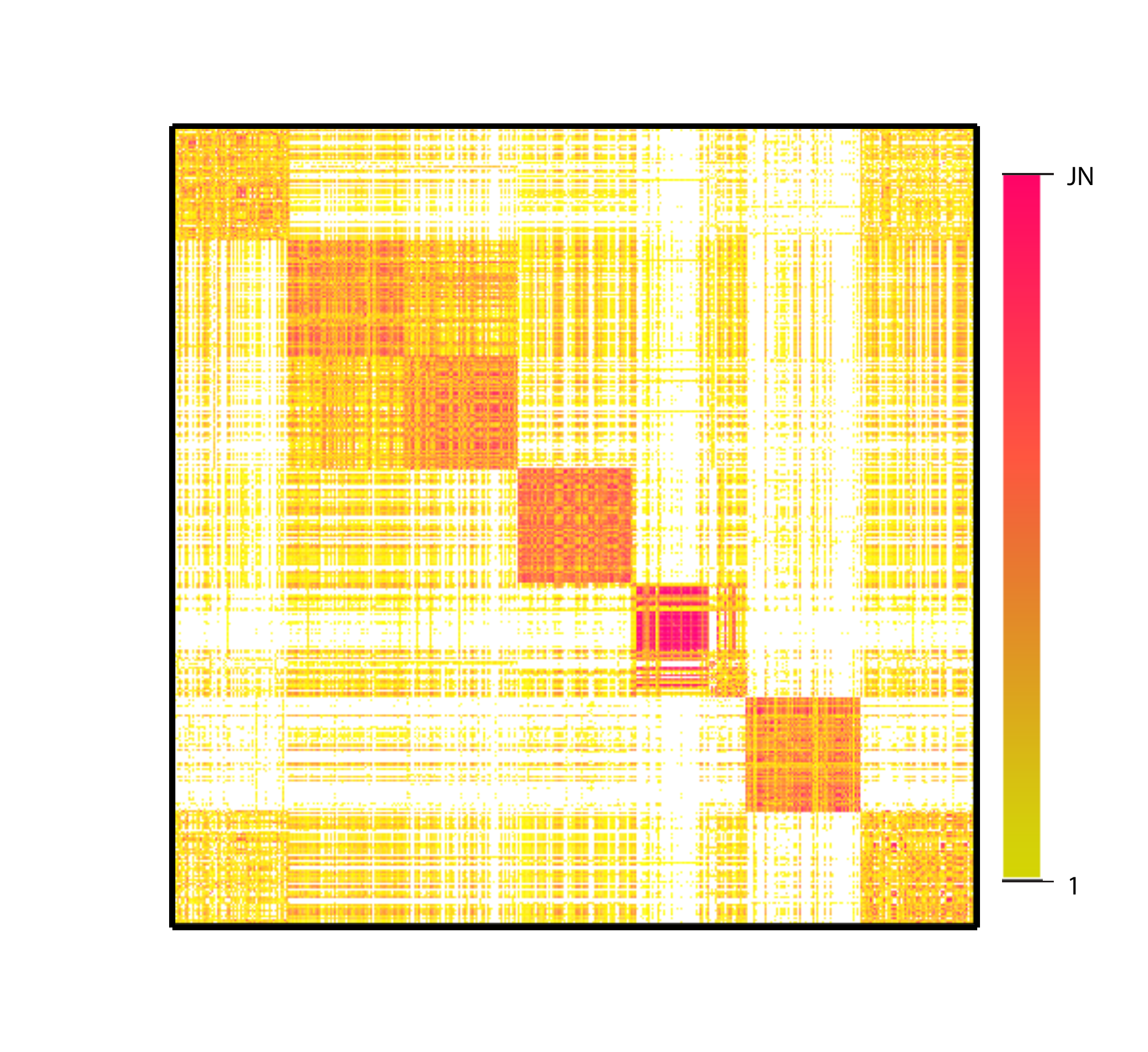}
\caption{Consensus Matrix prior to iteration}
\end{subfigure}

\begin{subfigure}[h!]{\linewidth}
\centering
\includegraphics[scale=.35]{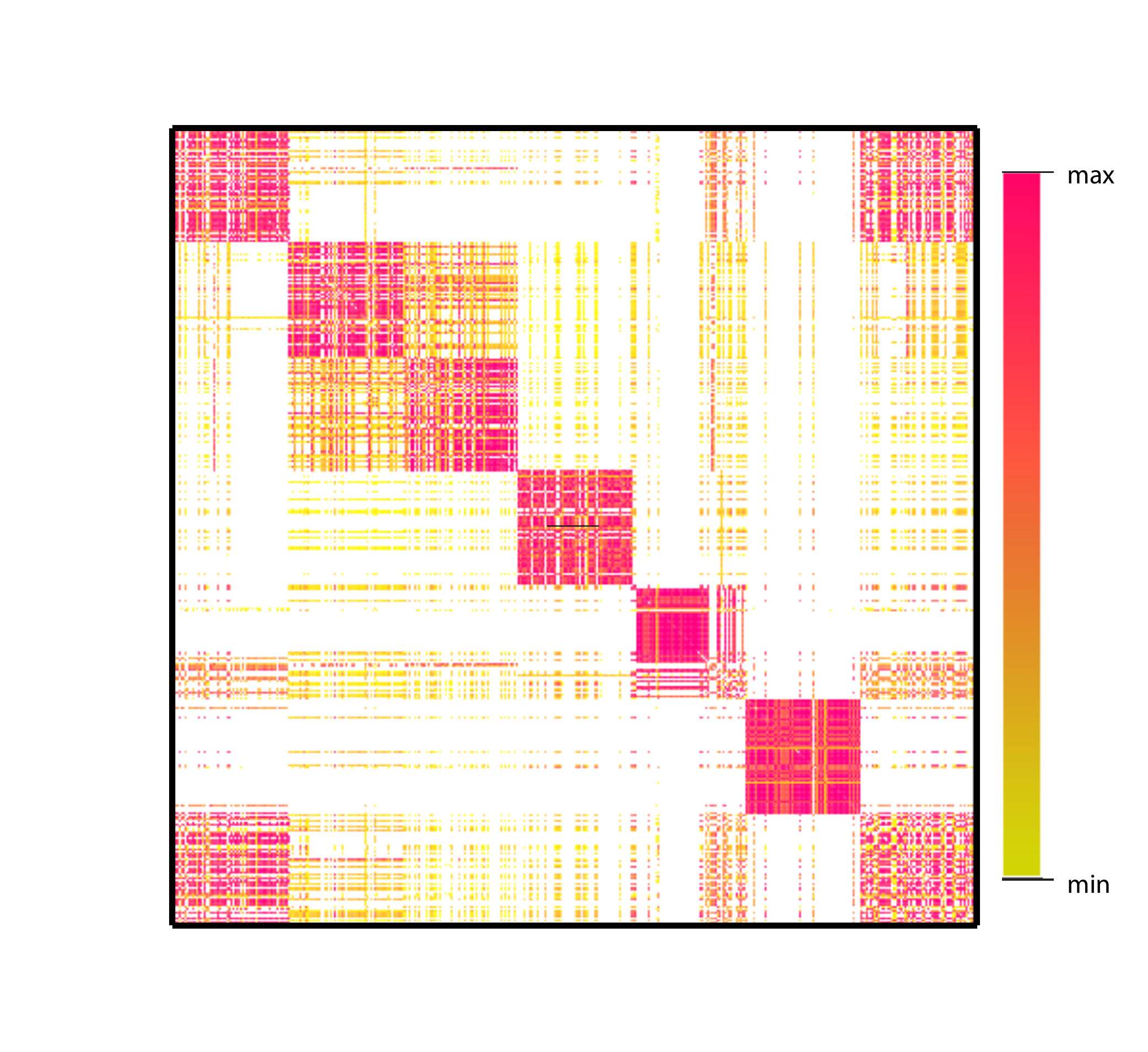}
\caption{Consensus Matrix after 2 iterations}
\end{subfigure}
\caption{The Uncoupling Effect of Iteration}
\label{fig:ngconsensus}
\end{figure}

 The documents were clustered using $\tilde{k}=[10,11,\dots,20]$ clusters and a drop tolerance of $\tau=0.1$. David Gleich's VISMATRIX tool allows us to visualize our matrices as heat maps. In Figure \ref{fig:ngconsensus}, observe the difference between the consensus matrix prior to iteration (after the drop tolerance enforced) and after just two iterations. Each non-zero entry in the matrix is represented by a colored pixel. The colorbar on the right indicates the magnitude of the entries by color. \par

\begin{figure}[h!]
\begin{subfigure}[h]{\linewidth}
\centering
\includegraphics[scale=.35]{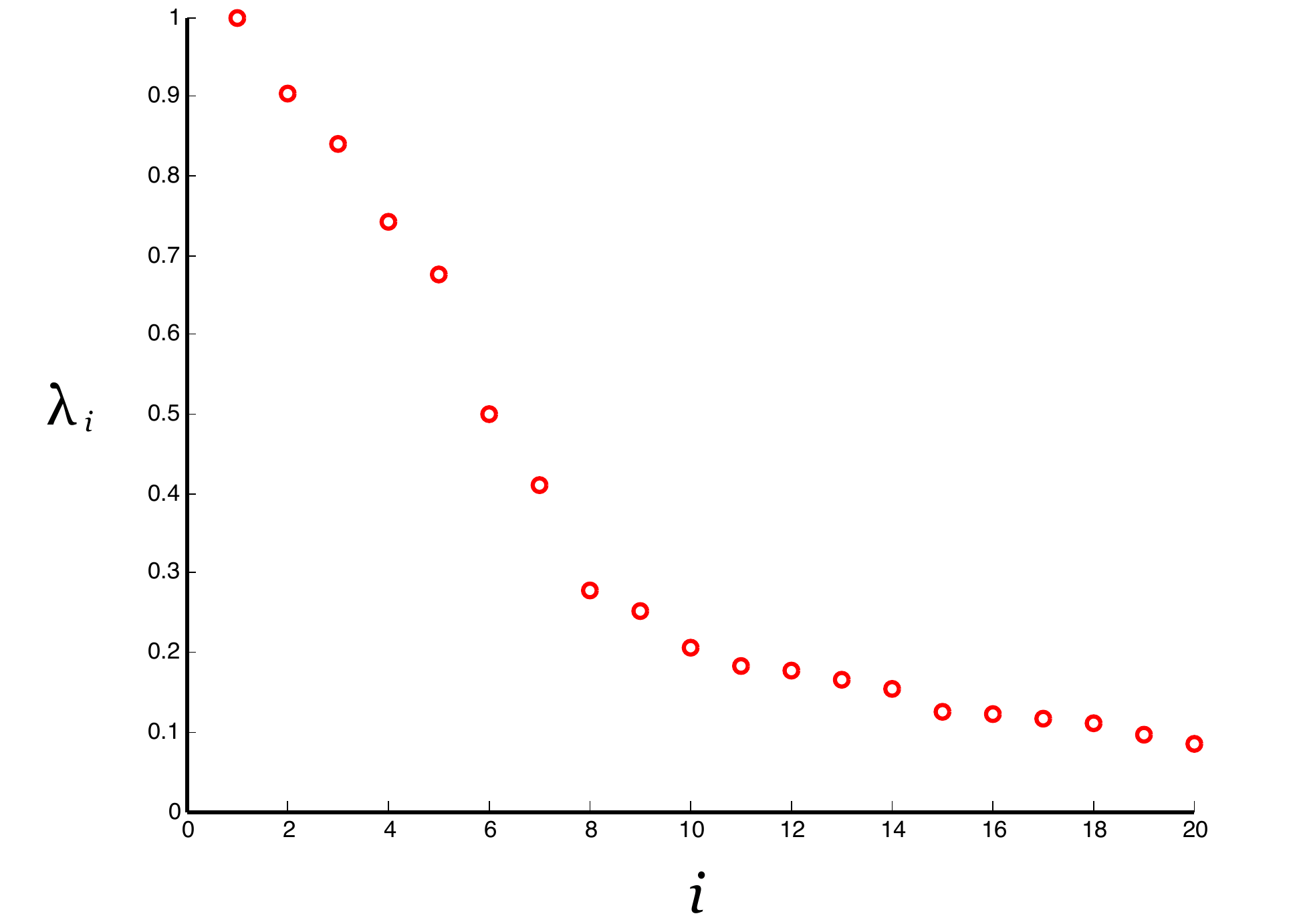}
\caption{Eigenvalues prior to iteration}
\end{subfigure}

\begin{subfigure}[h!]{\linewidth}
\centering
\includegraphics[scale=.35]{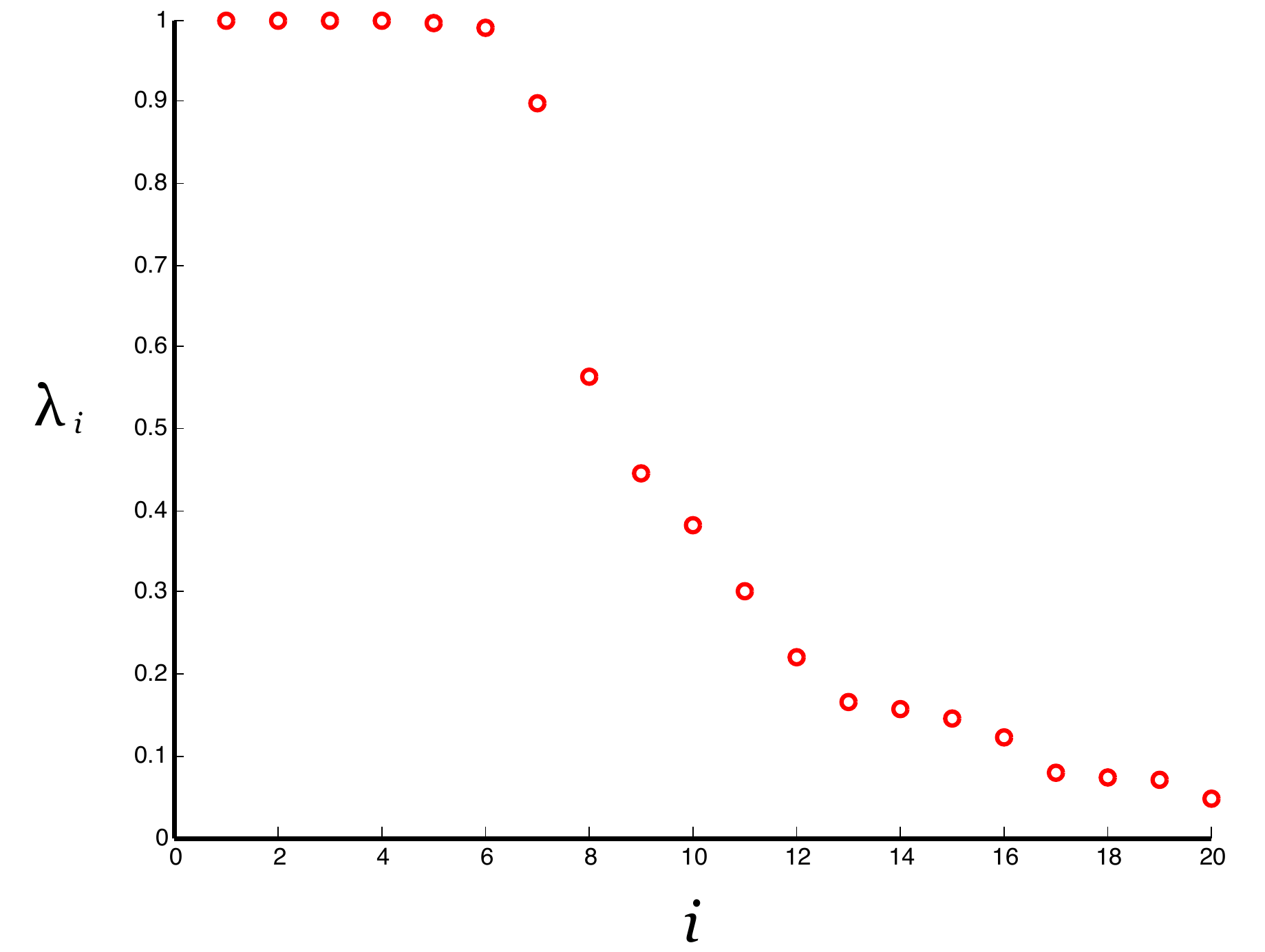}
\caption{Eigenvalues after 2 iterations}
\end{subfigure}
\caption{Newsgroups Dataset - ICC Results}
\label{fig:ngeigs}
\end{figure}

After 2 iterations, the magnitude of intra-cluster similarites are clearly larger and the magnitudes of inter-cluster similarities (noise) are noticeably diminished. Note the strong similarities between clusters 1 and 7, and some weaker similarities between clusters 2 and 3. This is due to the meaningful subcluster structure of the document collection: the categorical topics for clusters 1 and 7 are ``atheism'' and ``misc. religion" respectively and the topics for clusters 2 and 3 are ``computers- graphics'' and ``computers- OS MS windows misc''.  In fact, one of the beautiful aspects of our algorithm is its ability to detect this ``subcluster'' structure of data. \par
In Figure \ref{fig:ngeigs} we visualize the uncoupling effect of iteration by observing the difference in the eigenvalues of the transition probability matrix. Prior to iteration, the Perron cluster of eigenvalues is not apparent because there is still too much inter-cluster noise in the matrix. However, after 2 iterations, the Perron cluster is clearly visible, and contains the ``correct" number of $k=7$ eigenvalues. Furthermore, the $7^{th}$ eigenvalue belonging to the Perron-cluster is smaller in magnitude ($\lambda_7=0.89$) than the others ($\lambda_6=0.99$). This type of effect in the eigenvalues should cause the user to consider a subclustering situation like the one caused by the topics labels ``atheism'' and ``misc. religion" where one topic could clearly be considered a subtopic of another. \par 

\begin{figure}[h!]
\begin{subfigure}[h!]{\linewidth}
\centering
\includegraphics[scale=.35]{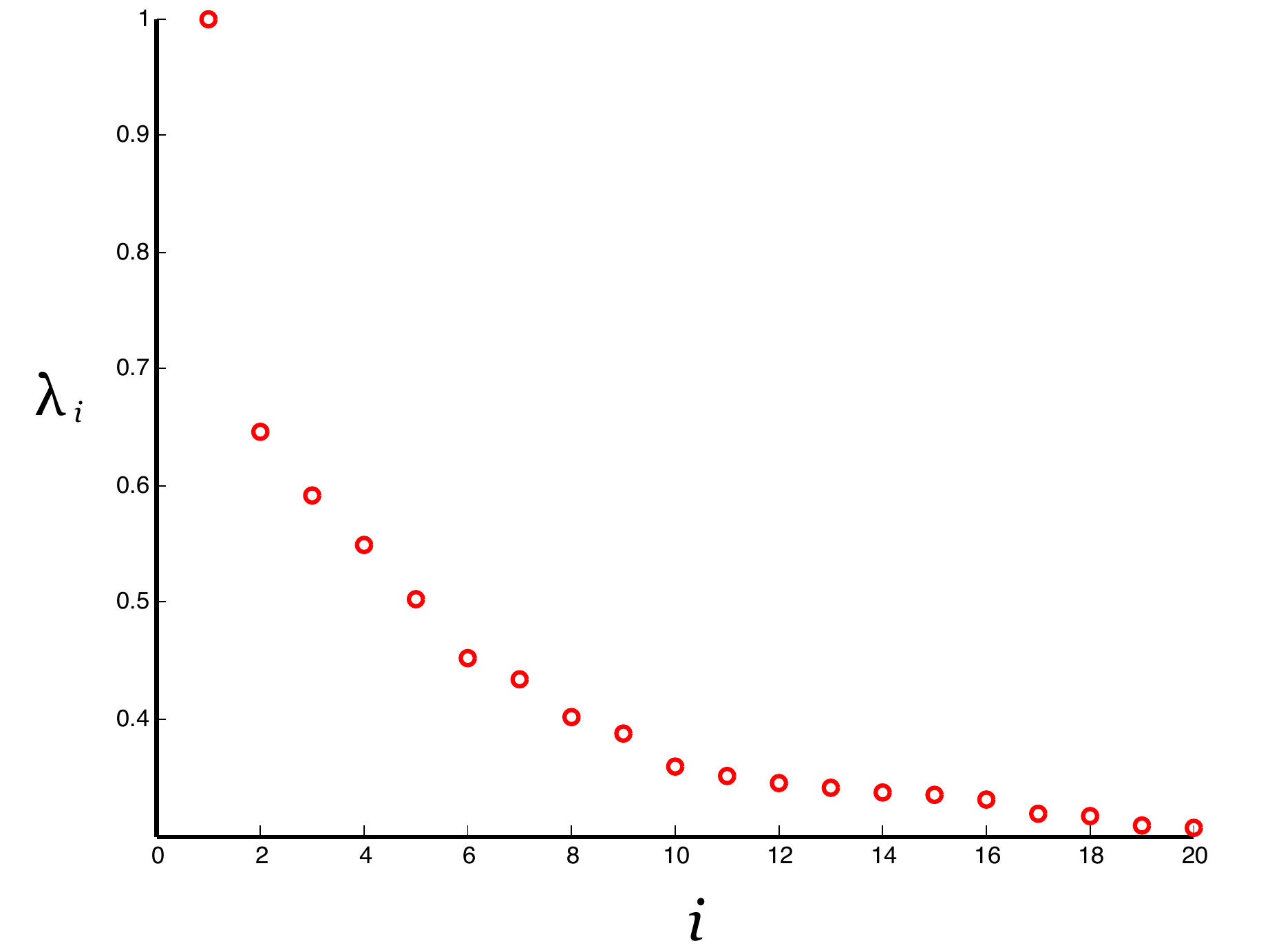}
\caption{Eigenvalues from Cosine Graph}
\end{subfigure}

\begin{subfigure}[h!]{\linewidth}
\centering
\includegraphics[scale=.35]{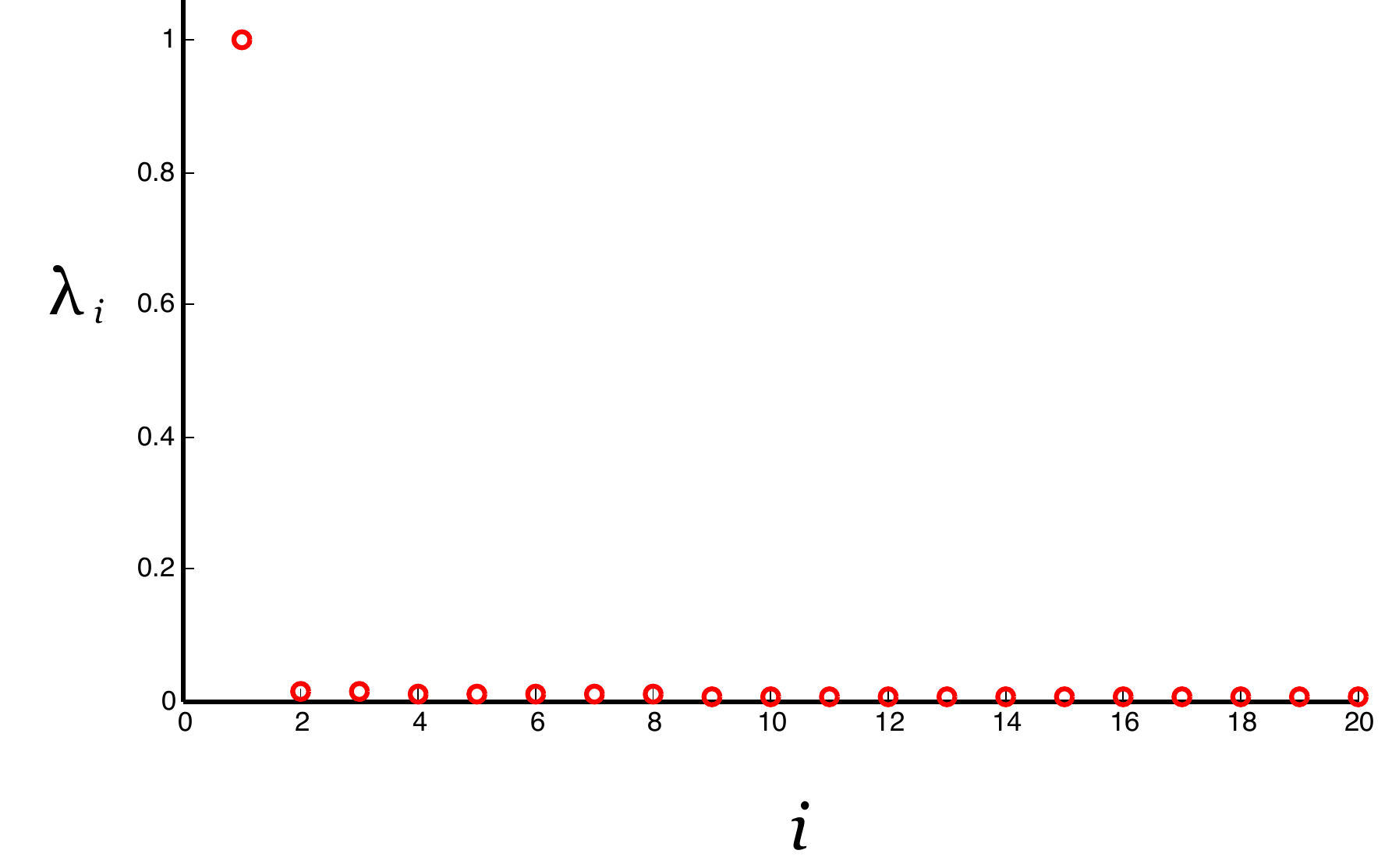}
\caption{Eigenvalues from Gaussian Graph}
\end{subfigure}

\caption{Newsgroups dataset}
\label{fig:NG7othersims}
\end{figure}

To complete our discussion of this text dataset, we present in Figure \ref{fig:NG7othersims} the eigenvalue plots of the Markov chains induced by two other similarity matrices, the cosine matrix and the Gaussian matrix. It is evident from this illustration that these measures of similarity fail to yield a nearly uncoupled Markov chain. 

\subsection{PenDigits17}
\begin{figure}[h!]

\begin{subfigure}[h]{\linewidth}
\centering
\includegraphics[scale=.35]{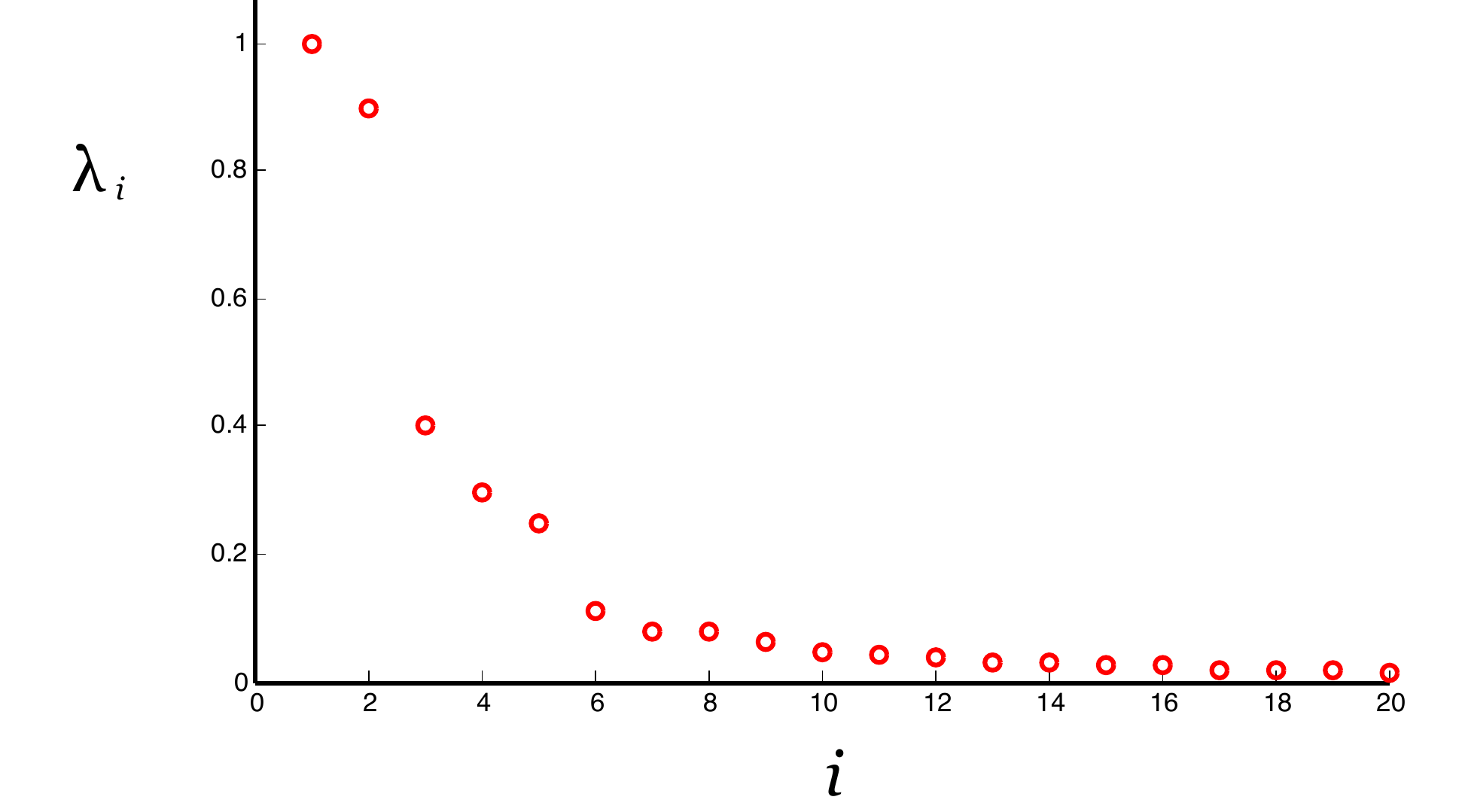}
\caption{Eigenvalues prior to iteration}
\end{subfigure}

\begin{subfigure}[h]{\linewidth}
\centering
\includegraphics[scale=.35]{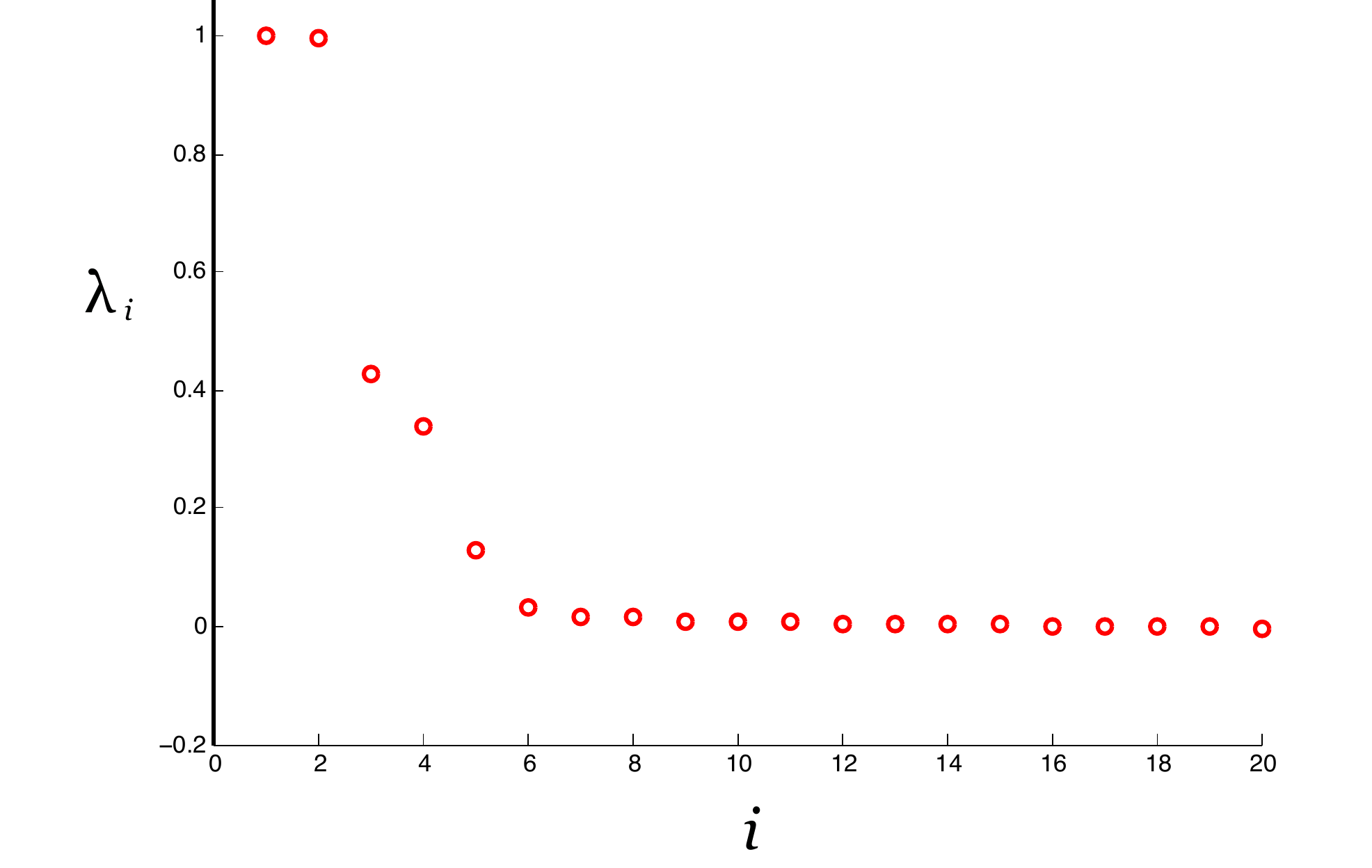}
\caption{Eigenvalues after 6 iterations}
\end{subfigure}

\caption{PenDigits17 Dataset - ICC Results}
\label{fig:pen17}
\end{figure}

PenDigits17 is a dataset, a subset of which was used in \cite{poweriteration}, which consists of coordinate observations made on handwritten digits. There are roughly 1000 instances each of $k=2$ digits, `1's and `7's, drawn by 44 writers. This is considered a difficult dataset because of the similarity of the two digits and the number of ways to draw each. The complete PenDigits dataset is available from the UCI machine learning repositiory \cite{uci}. For our experiments we used the sequence $\tilde{k}=[3,4,5,6]$ and a drop-tolerance $\tau=0.1$.  As seen in Figure \ref{fig:pen17} the Perron-cluster is convincing prior to iteration, and the system is almost completely uncoupled after 6 iterations.
\begin{figure}[ht!]
\begin{subfigure}[h]{\linewidth}
\centering
\includegraphics[scale=.35]{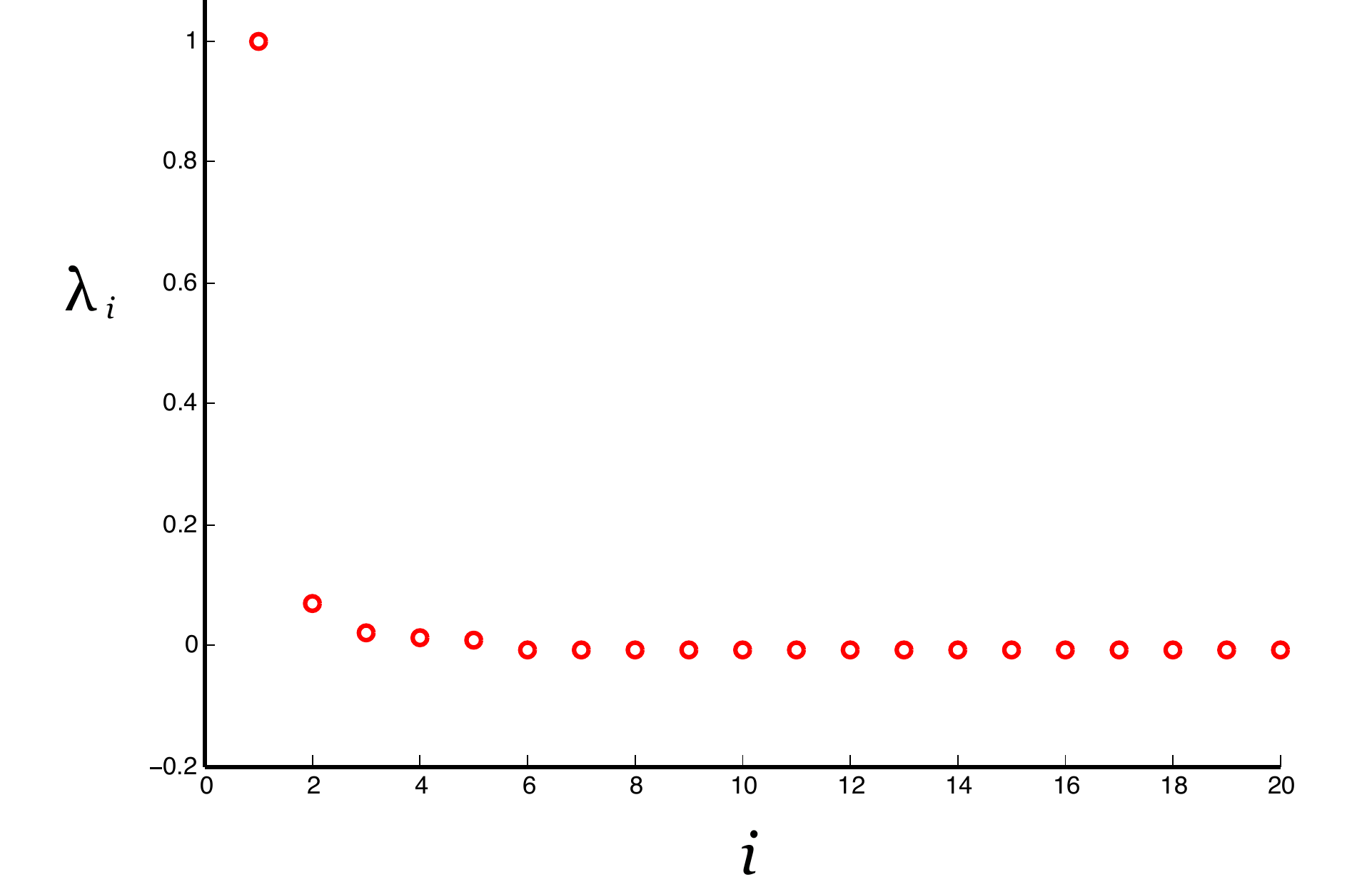}
\caption{Eigenvalues from Cosine Graph}
\end{subfigure}

\begin{subfigure}[ht]{\linewidth}
\centering
\includegraphics[scale=.35]{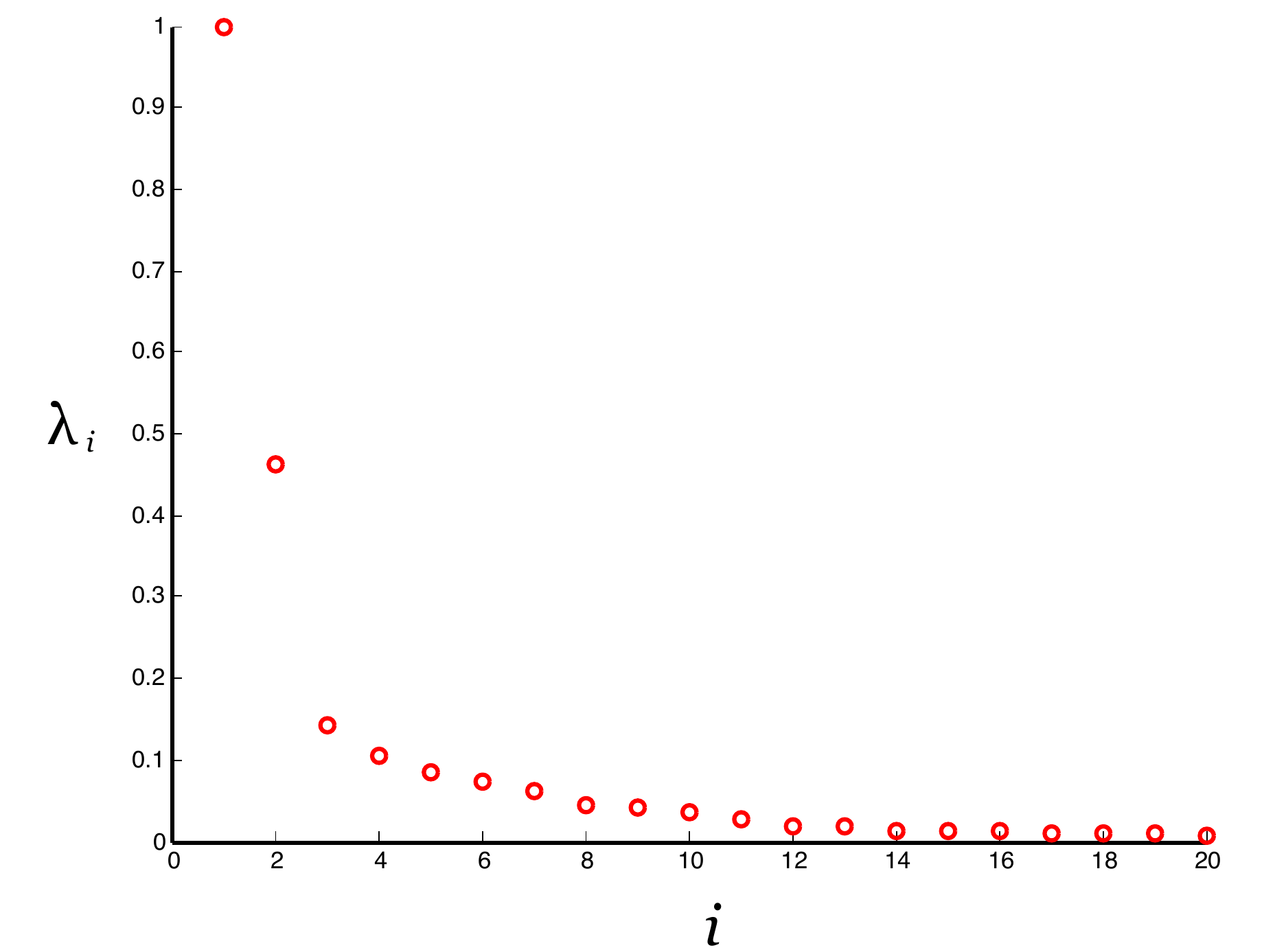}
\caption{Eigenvalues from Gaussian Graph}
\end{subfigure}
\caption{PenDigits17 Dataset}
\label{fig:pen17othersims}
\end{figure}
\par For the purposes of comparison, we observe the eigenvalues of the transition probability matrices associated with the graphs defined by the cosine similarity matrix (used for clustering of this dataset in \cite{poweriteration}) and the Gaussian similarity matrix. In Figure \ref{fig:pen17othersims} it is again clear that these similarity matrices are inadequate for determining the number of clusters.

\subsection{AGblog} 
\begin{figure}[h!]

\begin{subfigure}[h!]{\linewidth}

\includegraphics[scale=.35]{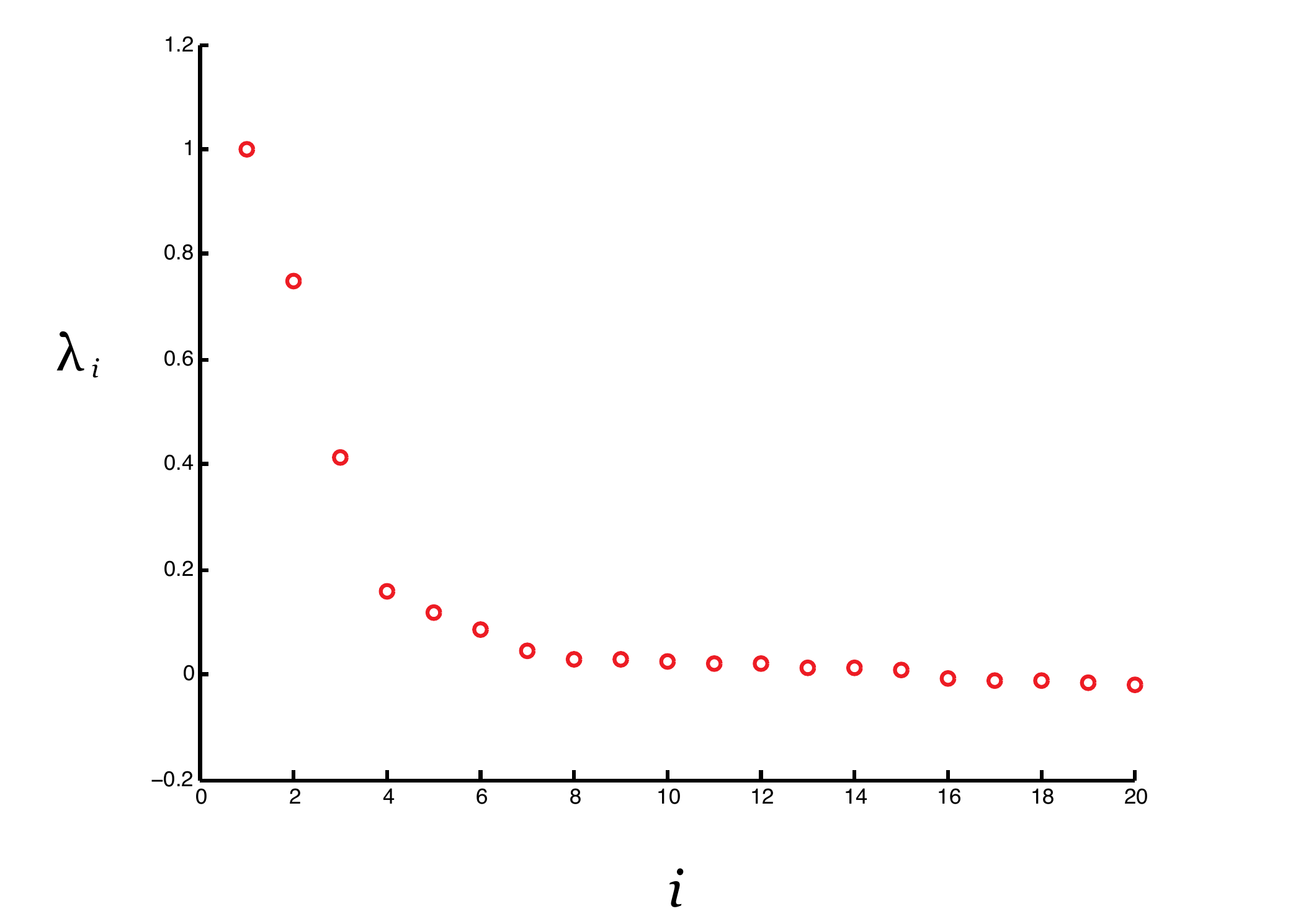}
\caption{Eigenvalues prior to iteration}
\end{subfigure}
%

\begin{subfigure}[h!]{\linewidth}
\centering
\includegraphics[scale=.35]{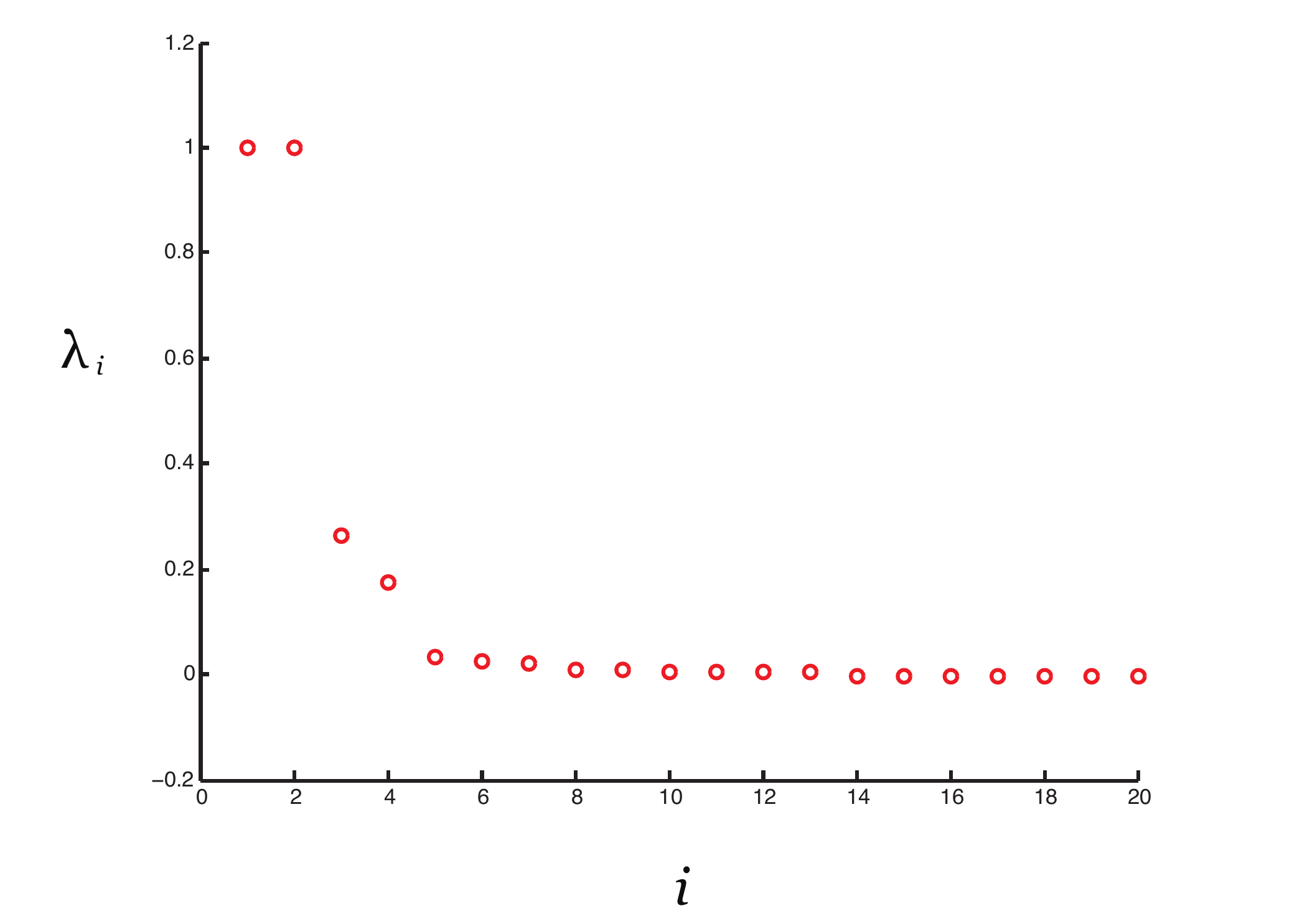}
\caption{Eigenvalues after 6 iterations}
\end{subfigure}
\caption{AGBlog Dataset - ICC Results}
\label{fig:agblog}
\end{figure}

is an undirected hyperlink network mined from 1222 political blogs. This dataset was used in \cite{poweriteration} and is described in \cite{agblog}. It contains $k=2$ clusters pertaining to the liberal and conservative division. We set our algorithms to find $\tilde{k}=[2:7]$ clusters with a drop tolerance of $\tau=0.2$. The resulting eigenvalue plots are displayed in Figure \ref{fig:agblog}. \par
\begin{figure}[h!]
\centering
\begin{subfigure}[h!]{\linewidth}
\includegraphics[scale=.35]{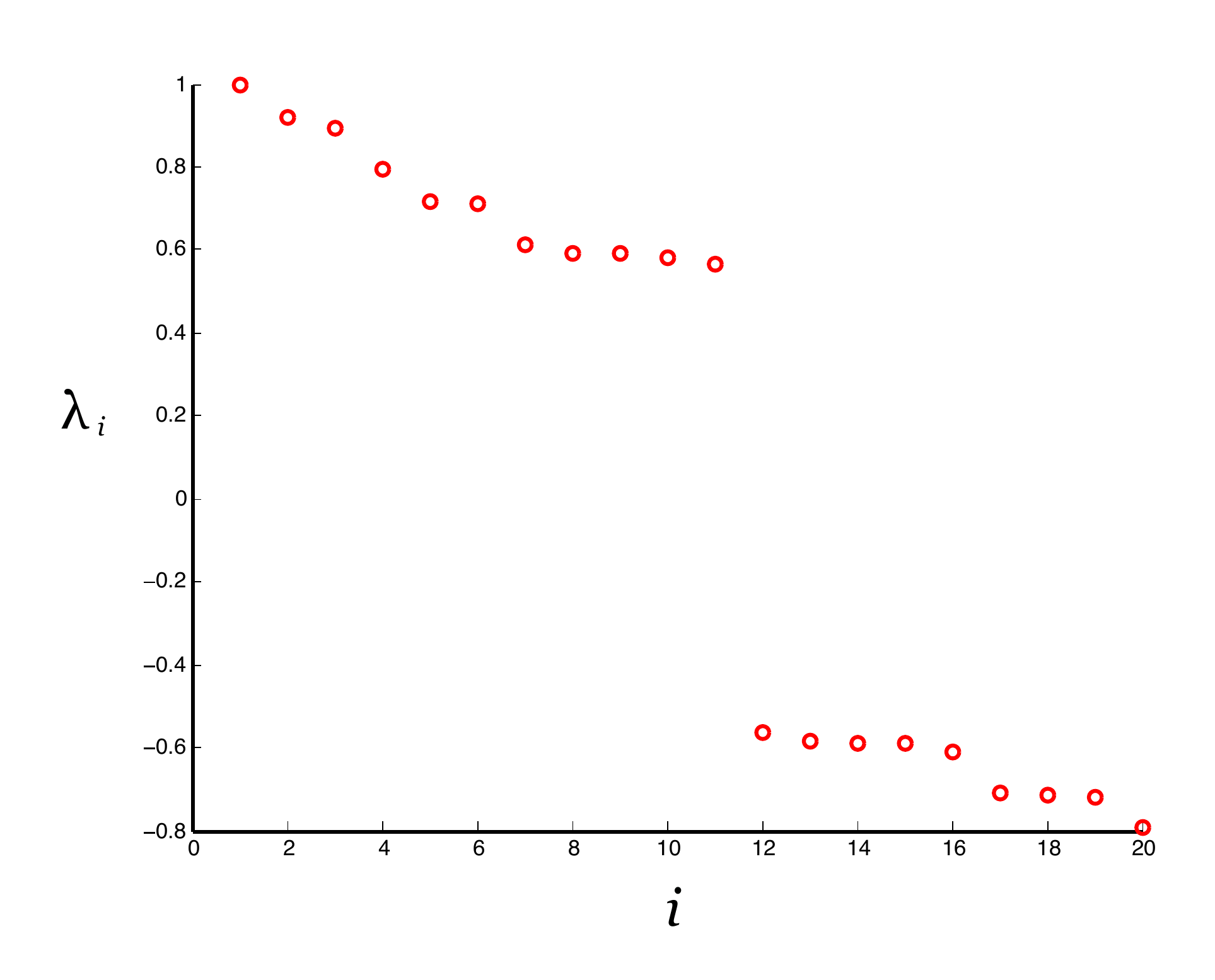}
\caption{Eigenvalues from Hyperlink Graph}
\end{subfigure}
\caption{AGBlog Dataset}
\label{fig:agblogaffinity}
\end{figure}

Figure \ref{fig:agblogaffinity} displays the eigenvalues from the Markov chain imposed by the similarity matrix used in \cite{poweriteration}, which was simply the original hyperlink matrix. This plot is particularly interesting because it does contain what appears to be a Perron cluster as defined by the large gap after the $11^{th}$ eigenvalue. There is, however, no indication in any type of analysis that suggests there are 11 communities in this dataset. We encourage readers to check out the visualizations of this graph at www4.ncsu.edu/\raise.17ex\hbox{$\scriptstyle\sim$}slrace which support this hypothesis.   We believe this ``uncoupling" is due, in fact, to rather isolated blogs that do not link to other sites with the frequency that others do. This example warns that outliers can have a misleading effect on the eigenvalues of an affinity matrix. 
\par It may be unreasonable to expect a graph clustering algorithm like \textit{Normalized Cut} (NCut) \cite{shi}, \textit{Power Iteration} (PIC) \cite{poweriteration}, or that of Ng, Jordan and Weiss (NJW) \cite{ng} to accurately divide a graph into two clusters when the eigenvalues of the associated Markov chain indicate eleven potential groups to be formed rather than two. In fact, we see the accuracy or purity of the clusters found by these algorithms may increase dramatically when the consensus matrix is used in place of the original matrix. Table \ref{tab:agblog} shows the purity of the clusterings found by these three spectral algorithms which were compared in \cite{poweriteration}. The consensus matrix makes the true clusters more obvious to the first two algorithms, and has little to no effect on the third. The cluster solutions from the three algorithms on the consensus matrix are identical. This type of agreement is important in practice because it gives the user an additional level of confidence in the cluster solution \cite{phdthesis}.\par
\begin{table}[h!]

    \begin{tabular}{|l|c|c|c|}
        \hline
        Similarity Matrix    & NCut & NJW  & PIC  \\ \hline
        Undirected Hyperlink & 0.52 & 0.52 & 0.96 \\ 
        Consensus Matrix     & 0.95 & 0.95 & 0.95 \\
        \hline
    \end{tabular}
\caption[centerlast]{Comparison of purity measurements for spectral algorithms on two similarity matrices for AGBlog Data}
\label{tab:agblog}
\end{table}

\section{Conclusions}
This paper demonstrates the effectiveness of Iterated Consensus Clustering (ICC) at the task of determining the number of clusters, $k$, in a dataset. Our main contribution is the formation of a consensus matrix from multiple algorithms and dimension reductions without prior knowledge of $k$. This consensus matrix is superior to other similarity matrices for determining $k$ due to the nearly uncoupled structure of its associated graph. If the graph of the initial consensus matrix is not nearly uncoupled, then the adjustments of iteration and drop tolerance outlined in Section \ref{sec:adjustments} will encourage such a structure.\par 
 Once the number of clusters is known, ICC has been previously been shown to obtain excellent clustering results by encouraging the underlying algorithms to agree upon a common solution through iteration \cite{thesis}. Here we have demonstrated that using the consensus similarity matrix instead of existing similarity matrices can improve the performance of existing spectral clustering algorithms as suggested in \cite{consensusspectral}. 

ICC is a flexible, exploratory method for determining the number of clusters. Its framework can be adapted to use any clustering algorithms or dimension reductions preferred by the user. This flexibility allows for scalability, given that the computation time of our method is dependent only upon the computation time of the algorithms used. The drop tolerance, $\tau$ can be changed to reflect the confidence the user has with their chosen clustering algorithms based upon the level of noise in the data. The range of values specified for $\tilde{k}$ and the level of dimension reduction (if any) can also changed for the purposes of investigation. 


%

\end{document}